\documentclass[10pt,twocolumn,letterpaper]{article}

\usepackage{iccv}
\usepackage{times}
\usepackage{epsfig}
\usepackage{graphicx}
\usepackage{amsmath}
\usepackage{amssymb}
\usepackage{pgf}
\usepackage{tikz}
\usepackage{caption}
\usepackage{subcaption}

\usepackage[pagebackref=true,breaklinks=true,letterpaper=true,colorlinks,bookmarks=false]{hyperref}

\iccvfinalcopy 

\def\iccvPaperID{12405} 

\begin{document}

\title{RelVAE: Generative Pretraining for few-shot Visual Relationship Detection}

\author{Sotiris Karapiperis\thanks{Work done during an internship at DeepLab.}\\
National Technical University of Athens\\
{\tt\small karapsot@gmail.com}
\and
Markos Diomataris\\
Max Planck Institute for Intelligent Systems\\
{\tt\small markos.diomataris@tuebingen.mpg.de}
\and
Vassilis Pitsikalis\\
DeepLab\\
{\tt\small vpitsik@deeplab.ai}
}

\maketitle

\begin{abstract}
Visual relations are complex, multimodal concepts that play an important role in the way humans perceive the world.
As a result of their complexity, high-quality, diverse and large scale datasets for visual relations are still absent.
In an attempt to overcome this data barrier, we choose to focus on the problem of few-shot Visual Relationship Detection (VRD), a setting that has been so far neglected by the community.
In this work we present the first pretraining method for few-shot predicate classification that does not require any annotated relations.
We achieve this by introducing a generative model that is able to capture the variation of semantic, visual and spatial information of relations inside a latent space and later exploiting its representations in order to achieve efficient few-shot classification.
We construct few-shot training splits and show quantitative experiments on VG200 and VRD datasets where our model outperforms the baselines. Lastly we attempt to interpret the decisions of the model by conducting various qualitative experiments.
\end{abstract}

\section{Introduction}
Visual relations are concepts able to encode a combination of high-level semantics together with low level spatial information.
For example, by knowing that a person is sitting on a chair next to a table we can assume that the action of "eating" might follow, but we are also able to constrain the human's possible pose and spatial configuration of the scene.
Recently, it has been shown that even though the scale and training data of vision-language models has increased dramatically \cite{radford2021}, interestingly, their ability to understand and represent visual relations is still poor \cite{parcalabescu_2022}.
Visual Relationship Detection (VRD) is the task of detecting the objects in a given image, classifying them into the available categories and then classifying the interaction type as one of the available predicates.
Typically the relations are represented as triplets $<$subject,predicate,object$>$.
VRD bridges the gap between low level tasks (\eg object detection) and high level understanding tasks (\eg image captioning, Visual Question Answering).

One of the most notorious problems that all commonly used datasets~\cite{Krishna_2017,Lu_2016} suffer from is the long tail distribution regarding the predicates. This is not only caused by the fact that some predicates naturally occur more frequently than others but also by the human nature of annotator's biases that prefer to annotate the most common predicates in an image that contains other predicates too ~\cite{Suhail21}. Most recent approaches propose frameworks that attempt to remove the bias from models trained with these datasets. Besides the aforementioned drawback, it is quite common that the annotated relations contain errors injecting noise into the training process. In this work we take a step back and argue that investigating the problem under the few-shot regime is of great importance.

\begin{figure}[t]
\centering
\includegraphics[width=1\linewidth]{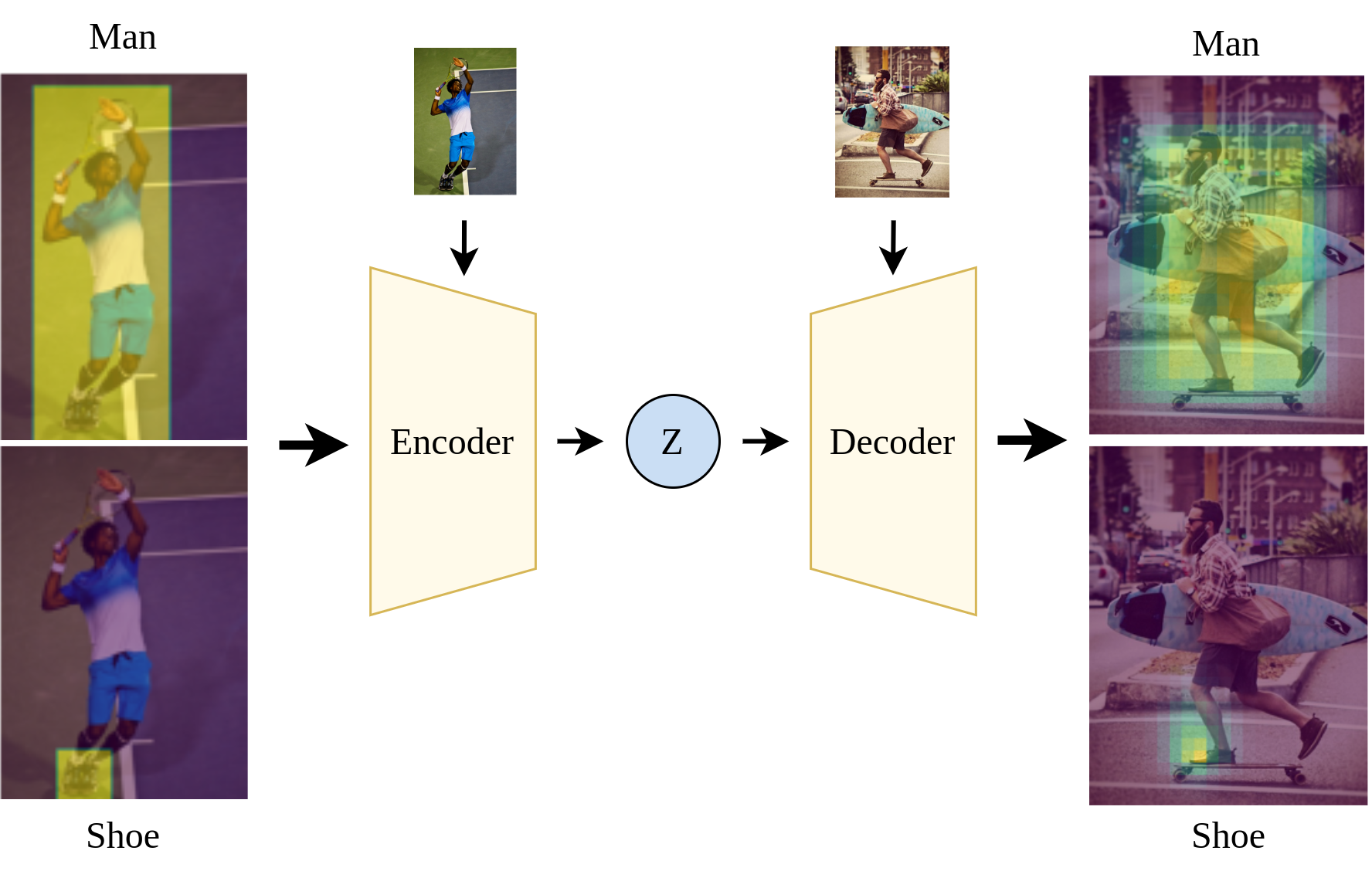}
\caption{RelVAE creates a latent space able to encode the semantic,visual and spatial variation of the context of relations . In this figure we showcase the ability of the model to produce meaningfull latent features that when decoded on a new image produce a similar context.}
\label{fig:teaser}
\end{figure}


Being able to learn a relation from just a few examples has numerous advantages over models that need to be trained on large scale VRD datasets. It can help avoid the bias of models trained on long tail datasets since we can easily build small balanced datasets that are also easy to check for wrong annotations. Furthermore learning a new relation requires the user to just annotate a few examples.
Training directly a predicate classifier with $k$ examples,$k \in {1,2,5}$, introduces new challenges. Each predicate can occur with a variety of context, semantics, visual and spatial configuration. Capturing this variance with as few as $k$ examples is not possible and finding a pretraining scheme that can capture this and produce features that transfer knowledge between different contexts is not trivial. Few-shot classification has been extensively studied in tasks such as image classification ~\cite{Snell2017,Vinyals2016}, but no one yet to the best of our knowledge has tackled VRD as a truly few-shot task i.e. pretrain without using any annotations.

The $<$subject,object$>$ pair of a triplet, including their semantics, visual and spatial information, is commonly called the $context$ of the relation. Previous works have already shown the importance of context towards identifying a relation ~\cite{Zellers_2017,Gkanatsios_2019b,Zhuang_2017}. We argue that the contexts surrounding some types of predicates are similar and that we can exploit that pattern in order to achieve few-shot classification. More specifically if we create a feature space where similar contexts are close between them, then we can use these features as the input to the few-shot classifiers allowing them to predict the same predicate to similar contexts. These observations led us to the pretraining sheme introduced below.

VAEs~\cite{Kingma_2014,rezende14} are a common method to create a latent space where feature similarity translates to euclidean distance between latent codes. Thus, we use a conditional VAE that takes as input a context and attempts to encode in the latent space the semantic and spatial data of it by conditioning on the image. We train the model using contexts from the VG200 dataset \emph{without using} any predicate annotations. To investigate the effects of this pretraining on few-shot predicate classification we create few-shot training splits by selecting a subset of the VG200 and VRD datasets. We pass each context through the encoder and use the latent code as it's representation to train $k$-shot classifiers with $k \in \{1,2,5\}$ both on the VG200 and VRD datasets.
Through qualitative and quantitative experiments we show that the latent space is indeed able to encode semantic and spatial information and that this kind of pretraining creates features that are able to transfer knowledge across relations thus helping to achieve few-shot classification.
To the best of our knowledge this is the first attempt for a pretraining method that benefits few-shot visual relation learning that is scalable and dataset agnostic. Code will be available as supplementary material and publicly available upon publication.


\begin{figure}[t]

\tikzset{every picture/.style={line width=0.75pt}} 

\begin{tikzpicture}[x=0.75pt,y=0.75pt,yscale=-0.9,xscale=0.9]

\draw  [fill={rgb, 255:red, 202; green, 222; blue, 244 }  ,fill opacity=1 ] (235.75,150.29) .. controls (235.75,140.57) and (243.72,132.69) .. (253.56,132.69) .. controls (263.39,132.69) and (271.36,140.57) .. (271.36,150.29) .. controls (271.36,160.01) and (263.39,167.89) .. (253.56,167.89) .. controls (243.72,167.89) and (235.75,160.01) .. (235.75,150.29) -- cycle ;
\draw  [fill={rgb, 255:red, 255; green, 250; blue, 234 }  ,fill opacity=1 ] (185.18,125.65) -- (213.67,134.19) -- (213.67,166.39) -- (185.18,174.94) -- cycle ;
\draw  [fill={rgb, 255:red, 255; green, 250; blue, 234 }  ,fill opacity=1 ] (321.93,174.23) -- (293.44,165.68) -- (293.44,133.49) -- (321.93,124.94) -- cycle ;
\draw  [fill={rgb, 255:red, 227; green, 246; blue, 208 }  ,fill opacity=1 ] (235.04,71.43) -- (270.65,71.43) -- (270.65,106.64) -- (235.04,106.64) -- cycle ;
\draw  [fill={rgb, 255:red, 209; green, 206; blue, 206 } ,fill opacity=1 ] (87.29,116.39) -- (150.27,116.39) -- (150.27,127.76) -- (87.29,127.76) -- cycle ;
\draw  [fill={rgb, 255:red, 209; green, 206; blue, 206 }  ,fill opacity=1 ] (98.29,120.62) -- (161.27,120.62) -- (161.27,131.98) -- (98.29,131.98) -- cycle ;
\draw  [fill={rgb, 255:red, 209; green, 206; blue, 206 }  ,fill opacity=1 ] (99.71,137.62) -- (149.57,137.62) -- (149.57,165.78) -- (99.71,165.78) -- cycle ;
\draw  [fill={rgb, 255:red, 209; green, 206; blue, 206 }  ,fill opacity=1 ] (342.59,136.11) -- (407.03,136.11) -- (407.03,147.47) -- (342.59,147.47) -- cycle ;
\draw  [fill={rgb, 255:red, 209; green, 206; blue, 206 } ,fill opacity=1 ] (353.84,140.33) -- (418.29,140.33) -- (418.29,151.7) -- (353.84,151.7) -- cycle ;
\draw  [fill={rgb, 255:red, 209; green, 206; blue, 206 }  ,fill opacity=1 ] (153.13,141.14) -- (153.13,190.43) -- (124.64,190.43) -- (124.64,141.14) -- cycle ;
\draw    (199.73,93.16) -- (199.73,113) -- (199.73,117.62) ;
\draw [shift={(199.73,120.62)}, rotate = 270] [fill={rgb, 255:red, 0; green, 0; blue, 0 }  ][line width=0.08]  [draw opacity=0] (8.93,-4.29) -- (0,0) -- (8.93,4.29) -- cycle    ;
\draw    (199.73,93.16) -- (220.39,93.16) ;
\draw    (306.57,93.16) -- (306.57,117.62) ;
\draw [shift={(306.57,120.62)}, rotate = 270] [fill={rgb, 255:red, 0; green, 0; blue, 0 }  ][line width=0.08]  [draw opacity=0] (8.93,-4.29) -- (0,0) -- (8.93,4.29) -- cycle    ;
\draw    (306.57,93.16) -- (285.91,93.16) ;
\draw    (164.83,128.36) -- (171.83,132.73) -- (175.82,135.22) ;
\draw [shift={(178.36,136.81)}, rotate = 211.98] [fill={rgb, 255:red, 0; green, 0; blue, 0 }  ][line width=0.08]  [draw opacity=0] (8.93,-4.29) -- (0,0) -- (8.93,4.29) -- cycle    ;
\draw    (164.83,162.86) -- (174.08,158.56) -- (174.22,158.5) ;
\draw [shift={(176.94,157.23)}, rotate = 155.05] [fill={rgb, 255:red, 0; green, 0; blue, 0 }  ][line width=0.08]  [draw opacity=0] (8.93,-4.29) -- (0,0) -- (8.93,4.29) -- cycle    ;
\draw  [fill={rgb, 255:red, 209; green, 206; blue, 206 },fill opacity=1 ] (343.3,159.44) -- (393.15,159.44) -- (393.15,187.61) -- (343.3,187.61) -- cycle ;
\draw  [fill={rgb, 255:red, 209; green, 206; blue, 206 } ,fill opacity=1 ] (396.72,162.97) -- (396.72,212.25) -- (368.23,212.25) -- (368.23,162.97) -- cycle ;
\draw    (221.1,150.19) -- (225.67,150.19) -- (228.07,150.19) ;
\draw [shift={(231.07,150.19)}, rotate = 180] [fill={rgb, 255:red, 0; green, 0; blue, 0 }  ][line width=0.08]  [draw opacity=0] (8.93,-4.29) -- (0,0) -- (8.93,4.29) -- cycle    ;
\draw    (328.34,131.28) -- (336.15,125.85) ;
\draw [shift={(338.62,124.14)}, rotate = 145.2] [fill={rgb, 255:red, 0; green, 0; blue, 0 }  ][line width=0.08]  [draw opacity=0] (8.93,-4.29) -- (0,0) -- (8.93,4.29) -- cycle    ;
\draw    (327.22,162.86) -- (335.93,167.17) ;
\draw [shift={(338.62,168.5)}, rotate = 206.3] [fill={rgb, 255:red, 0; green, 0; blue, 0 }  ][line width=0.08]  [draw opacity=0] (8.93,-4.29) -- (0,0) -- (8.93,4.29) -- cycle    ;
\draw    (327.22,149.49) -- (335.8,146.3) ;
\draw [shift={(338.62,145.26)}, rotate = 159.66] [fill={rgb, 255:red, 0; green, 0; blue, 0 }  ][line width=0.08]  [draw opacity=0] (8.93,-4.29) -- (0,0) -- (8.93,4.29) -- cycle    ;
\draw    (277.36,149.49) -- (283,149.49) -- (284.34,149.49) ;
\draw [shift={(287.34,149.49)}, rotate = 180] [fill={rgb, 255:red, 0; green, 0; blue, 0 }  ][line width=0.08]  [draw opacity=0] (8.93,-4.29) -- (0,0) -- (8.93,4.29) -- cycle    ;
\draw    (252.44,195.26) -- (262.33,195.26) -- (263.68,195.26) ;
\draw [shift={(266.68,195.25)}, rotate = 179.99] [fill={rgb, 255:red, 0; green, 0; blue, 0 }  ][line width=0.08]  [draw opacity=0] (8.93,-4.29) -- (0,0) -- (8.93,4.29) -- cycle    ;
\draw    (252.44,195.26) -- (252.43,174.84) ;
\draw  [fill={rgb, 255:red, 247; green, 205; blue, 205 }  ,fill opacity=1 ] (290.18,170.71) -- (290.18,220) -- (272.07,220) -- (272.07,170.71) -- cycle ;
\draw  [fill={rgb, 255:red, 209; green, 206; blue, 206 }  ,fill opacity=1 ] (341.79,113.39) -- (404.77,113.39) -- (404.77,124.76) -- (341.79,124.76) -- cycle ;
\draw  [fill={rgb, 255:red, 209; green, 206; blue, 206 },fill opacity=1 ] (352.79,117.62) -- (415.77,117.62) -- (415.77,128.98) -- (352.79,128.98) -- cycle ;

\draw (101.68,120.25) node [anchor=north west][inner sep=0.75pt]  [font=\footnotesize] [align=left] {Semantics};
\draw (124.1,160.61) node [anchor=north west][inner sep=0.75pt]  [font=\footnotesize] [align=left] {BBox};
\draw (199,130) node [anchor=north west][inner sep=0.75pt]  [font=\footnotesize,rotate=270] [align=left] {Encoder};
\draw (322,129) node [anchor=north west][inner sep=0.75pt]  [font=\footnotesize,rotate=270] [align=left] {Decoder};
\draw (247.84,142.5) node [anchor=north west][inner sep=0.75pt]   [align=left] {Z};
\draw (235.49,83.12) node [anchor=north west][inner sep=0.75pt]  [font=\footnotesize] [align=left] {Image};
\draw (287,171) node [anchor=north west][inner sep=0.75pt]  [font=\footnotesize,rotate=270] [align=left] {Classifier};
\draw (355.53,140.07) node [anchor=north west][inner sep=0.75pt]  [font=\footnotesize] [align=left] {Visual Feat};
\draw (367.18,173.67) node [anchor=north west][inner sep=0.75pt]  [font=\footnotesize] [align=left] {Heat\\Maps};
\draw (356.18,117.25) node [anchor=north west][inner sep=0.75pt]  [font=\footnotesize] [align=left] {Semantics};

\end{tikzpicture}
\caption{High level architecture of RelVAE.}
\end{figure}
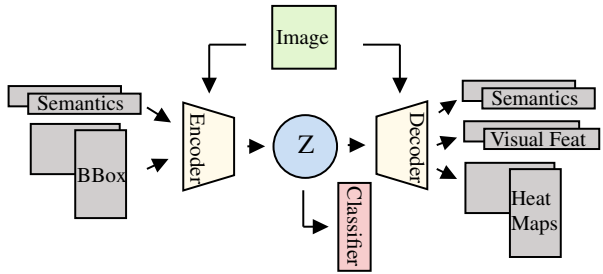

\section{Related Work}
\subsection{Visual Relationship Detection}
There are two common classification approaches on VRD: 1)~Treat every triplet $<$subject,predicate,object$>$ as a different class~\cite{Sadeghi11} 2)~Treat every predicate as a different class. The first approach suffers from scalability issues while the second struggles with the variability of the context surrounding a predicate \eg the predicate "wear" appears both in the $<man,shirt>$ and $<woman,jeans>$. In the seminal work of Lu et al.~\cite{Lu_2016}, the authors use the prior knowledge of word2vec to relate different contexts. Most current approaches create a complete graph between $n$ detected objects and make this graph sparse either by randomly sampling or by filtering edges with a network~\cite{Dai_2017,Yang18}.~\cite{Dai_2017} pass each resulting pair from a CRF inspired network to predict the predicate while ~\cite{Xu_2017,Yang18} use GNNs to aggregate context through message passing algorithms.~\cite{Zellers_2017} argue that there are Motifs between relationship co-occurances and attempt to exploit them.~\cite{Tang_2019} suggest that complete graphs are not efficient for message passing and proposes to dynamically compose binary tree structures. 

Later works attempt to construct model agnostic methods that can be used with an off-the-shelf model to intervene on its predictions or to further supervise the model in order to remove the bias due to the long tail distribution~\cite{Tang_2020,Suhail21,Diomataris21,Li_2022_CVPR,Li_2022_CVPR_2,Goel_2022_CVPR,Dong_2022_CVPR,Chen_2023_WACV}.~\cite{Tang_2020} attempts to remove the bias by looking at the causal graph between the Image, the Object detections, the Object labels and the predictions. Authors in~\cite{Li_2022_CVPR} introduce a loss that attempts to help the least common predicates while it does not decrease the performance on the most common ones and~\cite{Suhail21} proposes a loss that takes into account the structure of the output space (e.g. pairs of relations that cannot coexist). Lu et al.~\cite{Lu21} employ the same motivation but treat VRD as a seq2seq problem using transformers. The authors in~\cite{Diomataris21} attempt to improve the precision of a student model by supervising it with a teacher trained to produce image grounded relationships. Lastly ~\cite{Li_2022_CVPR_2,Goel_2022_CVPR,Chen_2023_WACV} attempt to correct the noisy labels, populate the dataset with implicit relations and incorporate external knowledge respectively.
\subsection{Few-shot VRD}
Event though the most commonly used SGG/VRD datasets follow a long tail distribution the few-shot setting of VRD has not been extensively studied.~\cite{Dornadula_2019} uses a GNN and attempts to learn functions to produce effective node representations that encode spatial and semantic information. They pretrain the model on the 25 most frequent predicates of VG200 and use the remaining 25 predicates to train k-shot classifiers.~\cite{Wang20} follows the most common formulation of few-shot where the model is given a Support set of N predicates with K examples each( e.g. N-way K-shot) and is evaluated on a Query set containing M examples with the same predicates. The authors use a dual GNN and attempt to predict the nodes of the Query set based on the nodes of the Support set. They create two new few-shot datasets by sampling from the original datasets. Our work differs in that we pretrain the model only on $<$subject,object$>$ pairs, without seeing any predicates, to create a meaningfull latent space and use K examples of all predicates(frequent and unfrequent) to train K-shot classifiers.


\begin{figure}[t]
\centering

\tikzset{every picture/.style={line width=0.75pt}} 

\begin{tikzpicture}[x=0.75pt,y=0.75pt,yscale=-0.9,xscale=0.9]

\draw  [color={rgb, 255:red, 255; green, 250; blue, 234 }  ,draw opacity=1 ][fill={rgb, 255:red, 255; green, 250; blue, 234 }  ,fill opacity=1] (345.75,268.23) .. controls (345.7,255.77) and (355.76,245.62) .. (368.22,245.55) -- (533.57,244.65) .. controls (546.03,244.59) and (556.17,254.63) .. (556.23,267.09) -- (556.51,334.76) .. controls (556.56,347.22) and (546.5,357.37) .. (534.05,357.44) -- (368.69,358.33) .. controls (356.23,358.4) and (346.09,348.36) .. (346.04,335.9) -- cycle ;
\draw  [fill={rgb, 255:red, 195; green, 214; blue, 252 }  ,fill opacity=1 ] (337.04,241.21) .. controls (337.04,242.78) and (335.77,244.06) .. (334.19,244.06) -- (299.89,244.06) .. controls (298.32,244.06) and (297.04,242.78) .. (297.05,241.21) -- (297.06,232.67) .. controls (297.06,231.1) and (298.34,229.83) .. (299.91,229.83) -- (334.21,229.83) .. controls (335.78,229.83) and (337.06,231.1) .. (337.05,232.67) -- cycle ;
\draw  [fill={rgb, 255:red, 209; green, 206; blue, 206 }  ,fill opacity=1 ] (228.03,232.14) -- (273.24,232.14) -- (273.2,243.89) -- (228,243.89) -- cycle ;
\draw  [fill={rgb, 255:red, 209; green, 206; blue, 206 }  ,fill opacity=1 ] (233.84,238.99) -- (279.04,238.99) -- (279.01,250.74) -- (233.81,250.74) -- cycle ;
\draw  [fill={rgb, 255:red, 203; green, 252; blue, 201 }  ,fill opacity=1 ] (379.32,238.03) -- (371.86,230.65) -- (371.83,221.9) -- (380.82,221.85) -- (388.27,229.24) -- (388.31,237.98) -- cycle ; \draw   (371.83,221.9) -- (379.28,229.29) -- (379.32,238.03) ; \draw   (379.28,229.29) -- (388.27,229.24) ;
\draw  [fill={rgb, 255:red, 203; green, 252; blue, 201 }  ,fill opacity=1 ] (370.44,238.08) -- (362.98,230.7) -- (362.95,221.95) -- (371.94,221.9) -- (379.39,229.28) -- (379.43,238.03) -- cycle ; \draw   (362.95,221.95) -- (370.4,229.33) -- (370.44,238.08) ; \draw   (370.4,229.33) -- (379.39,229.28) ;
\draw  [fill={rgb, 255:red, 203; green, 252; blue, 201 }  ,fill opacity=1 ] (361.56,238.13) -- (354.11,230.74) -- (354.07,222) -- (363.06,221.95) -- (370.52,229.33) -- (370.55,238.08) -- cycle ; \draw   (354.07,222) -- (361.52,229.38) -- (361.56,238.13) ; \draw   (361.52,229.38) -- (370.52,229.33) ;
\draw  [fill={rgb, 255:red, 203; green, 252; blue, 201 }  ,fill opacity=1 ] (379.28,228.95) -- (371.82,221.57) -- (371.79,212.82) -- (380.78,212.77) -- (388.23,220.16) -- (388.27,228.9) -- cycle ; \draw   (371.79,212.82) -- (379.24,220.21) -- (379.28,228.95) ; \draw   (379.24,220.21) -- (388.23,220.16) ;
\draw  [fill={rgb, 255:red, 203; green, 252; blue, 201 }  ,fill opacity=1 ] (370.4,229) -- (362.95,221.62) -- (362.91,212.87) -- (371.9,212.82) -- (379.36,220.2) -- (379.39,228.95) -- cycle ; \draw   (362.91,212.87) -- (370.36,220.25) -- (370.4,229) ; \draw   (370.36,220.25) -- (379.36,220.2) ;
\draw  [fill={rgb, 255:red, 203; green, 252; blue, 201 }  ,fill opacity=1 ] (361.52,229.05) -- (354.07,221.66) -- (354.03,212.92) -- (363.02,212.87) -- (370.48,220.25) -- (370.51,229) -- cycle ; \draw   (354.03,212.92) -- (361.49,220.3) -- (361.52,229.05) ; \draw   (361.49,220.3) -- (370.48,220.25) ;
\draw  [fill={rgb, 255:red, 203; green, 252; blue, 201 }  ,fill opacity=1 ] (379.24,219.62) -- (371.79,212.23) -- (371.75,203.49) -- (380.74,203.44) -- (388.19,210.82) -- (388.23,219.57) -- cycle ; \draw   (371.75,203.49) -- (379.2,210.87) -- (379.24,219.62) ; \draw   (379.2,210.87) -- (388.19,210.82) ;
\draw  [fill={rgb, 255:red, 203; green, 252; blue, 201 }  ,fill opacity=1 ] (370.36,219.66) -- (362.91,212.28) -- (362.87,203.54) -- (371.86,203.49) -- (379.32,210.87) -- (379.35,219.61) -- cycle ; \draw   (362.87,203.54) -- (370.33,210.92) -- (370.36,219.66) ; \draw   (370.33,210.92) -- (379.32,210.87) ;
\draw  [fill={rgb, 255:red, 203; green, 252; blue, 201 }  ,fill opacity=1 ] (361.48,219.71) -- (354.03,212.33) -- (353.99,203.58) -- (362.99,203.54) -- (370.44,210.92) -- (370.48,219.66) -- cycle ; \draw   (353.99,203.58) -- (361.45,210.97) -- (361.48,219.71) ; \draw   (361.45,210.97) -- (370.44,210.92) ;

\draw  [fill={rgb, 255:red, 195; green, 214; blue, 252 }  ,fill opacity=1 ] (297.67,261.15) .. controls (297.67,259.47) and (299.03,258.1) .. (300.71,258.1) -- (337.85,258.1) .. controls (339.53,258.1) and (340.9,259.47) .. (340.9,261.15) -- (340.91,270.28) .. controls (340.91,271.96) and (339.55,273.32) .. (337.87,273.32) -- (300.73,273.32) .. controls (299.05,273.32) and (297.68,271.96) .. (297.68,270.28) -- cycle ;
\draw  [fill={rgb, 255:red, 209; green, 206; blue, 206 }  ,fill opacity=1 ] (237.44,259.41) -- (268.08,259.25) -- (268.2,285.77) -- (237.55,285.94) -- cycle ;
\draw  [fill={rgb, 255:red, 209; green, 206; blue, 206 }  ,fill opacity=1 ] (271.28,262.6) -- (271.28,309.02) -- (246.58,309.05) -- (246.58,262.63) -- cycle ;
\draw  [fill={rgb, 255:red, 252; green, 205; blue, 205 }  ,fill opacity=1 ] (294.65,285.65) -- (339.74,285.65) -- (339.74,296.33) -- (294.65,296.33) -- cycle ;
\draw  [fill={rgb, 255:red, 252; green, 205; blue, 205 }  ,fill opacity=1 ] (299.61,289.52) -- (344.7,289.52) -- (344.7,300.21) -- (299.61,300.21) -- cycle ;
\draw  [fill={rgb, 255:red, 195; green, 214; blue, 252 }  ,fill opacity=1 ] (382.19,259.87) .. controls (385.22,259.87) and (387.67,262.33) .. (387.67,265.36) -- (387.67,308.71) .. controls (387.67,311.74) and (385.22,314.19) .. (382.19,314.19) -- (365.74,314.19) .. controls (362.72,314.19) and (360.26,311.74) .. (360.26,308.71) -- (360.26,265.36) .. controls (360.26,262.33) and (362.72,259.87) .. (365.74,259.87) -- cycle ;
\draw  [fill={rgb, 255:red, 252; green, 205; blue, 205 }  ,fill opacity=1 ] (404.28,285.51) -- (444.03,285.29) -- (444.08,296.91) -- (404.33,297.13) -- cycle ;
\draw  [fill={rgb, 255:red, 252; green, 205; blue, 205 }  ,fill opacity=1 ] (408.66,289.73) -- (448.41,289.52) -- (448.46,301.13) -- (408.71,301.35) -- cycle ;
\draw  [fill={rgb, 255:red, 195; green, 214; blue, 252 }  ,fill opacity=1 ] (472.23,284.43) .. controls (473.55,284.43) and (474.61,285.5) .. (474.61,286.82) -- (474.61,328.46) .. controls (474.61,329.78) and (473.55,330.85) .. (472.23,330.85) -- (465.07,330.85) .. controls (463.76,330.85) and (462.69,329.78) .. (462.69,328.46) -- (462.69,286.82) .. controls (462.69,285.5) and (463.76,284.43) .. (465.07,284.43) -- cycle ;
\draw  [fill={rgb, 255:red, 195; green, 214; blue, 252 }  ,fill opacity=1 ] (502.32,252.7) .. controls (503.43,252.7) and (504.34,253.6) .. (504.34,254.72) -- (504.34,294.46) .. controls (504.34,295.58) and (503.43,296.48) .. (502.32,296.48) -- (496.28,296.48) .. controls (495.16,296.48) and (494.26,295.58) .. (494.26,294.46) -- (494.26,254.72) .. controls (494.26,253.6) and (495.16,252.7) .. (496.28,252.7) -- cycle ;
\draw  [fill={rgb, 255:red, 195; green, 214; blue, 252 }  ,fill opacity=1 ] (503.03,303.21) .. controls (504.14,303.21) and (505.04,304.11) .. (505.04,305.23) -- (505.04,345.72) .. controls (505.04,346.83) and (504.14,347.74) .. (503.03,347.74) -- (496.98,347.74) .. controls (495.87,347.74) and (494.97,346.83) .. (494.97,345.72) -- (494.97,305.23) .. controls (494.97,304.11) and (495.87,303.21) .. (496.98,303.21) -- cycle ;
\draw  [fill={rgb, 255:red, 252; green, 205; blue, 205 }  ,fill opacity=1 ] (523.62,271.17) -- (548.42,271.17) -- (548.42,278.41) -- (523.62,278.41) -- cycle ;
\draw  [fill={rgb, 255:red, 252; green, 205; blue, 205 }  ,fill opacity=1 ] (523.3,314.67) -- (550.09,314.67) -- (550.09,322.96) -- (523.3,322.96) -- cycle ;
\draw    (282.22,238.41) -- (292.91,238.2) ;
\draw [shift={(295.91,238.15)}, rotate = 178.9] [fill={rgb, 255:red, 0; green, 0; blue, 0 }  ][line width=0.08]  [draw opacity=0] (7.14,-3.43) -- (0,0) -- (7.14,3.43) -- cycle    ;
\draw    (273.74,266.16) -- (290.56,266.26) ;
\draw [shift={(293.56,266.27)}, rotate = 180.34] [fill={rgb, 255:red, 0; green, 0; blue, 0 }  ][line width=0.08]  [draw opacity=0] (7.14,-3.43) -- (0,0) -- (7.14,3.43) -- cycle    ;
\draw    (317.7,216.47) -- (317.72,222.49) ;
\draw [shift={(317.74,225.49)}, rotate = 269.76] [fill={rgb, 255:red, 0; green, 0; blue, 0 }  ][line width=0.08]  [draw opacity=0] (7.14,-3.43) -- (0,0) -- (7.14,3.43) -- cycle    ;
\draw    (317.7,216.47) -- (351.48,216.54) ;
\draw    (374.56,241.29) -- (374.64,255.33) ;
\draw [shift={(374.65,258.33)}, rotate = 269.68] [fill={rgb, 255:red, 0; green, 0; blue, 0 }  ][line width=0.08]  [draw opacity=0] (7.14,-3.43) -- (0,0) -- (7.14,3.43) -- cycle    ;
\draw    (277.45,265.74) -- (277.08,325.87) ;
\draw    (284.87,238.4) -- (284.87,317.46) ;
\draw    (284.87,317.46) -- (456.44,317.46) ;
\draw [shift={(459.44,317.46)}, rotate = 180] [fill={rgb, 255:red, 0; green, 0; blue, 0 }  ][line width=0.08]  [draw opacity=0] (7.14,-3.43) -- (0,0) -- (7.14,3.43) -- cycle    ;
\draw    (277.08,325.87) -- (456.44,325.87) ;
\draw [shift={(459.44,325.87)}, rotate = 180] [fill={rgb, 255:red, 0; green, 0; blue, 0 }  ][line width=0.08]  [draw opacity=0] (7.14,-3.43) -- (0,0) -- (7.14,3.43) -- cycle    ;
\draw    (450.9,294.81) -- (456.3,294.78) ;
\draw [shift={(459.3,294.76)}, rotate = 179.69] [fill={rgb, 255:red, 0; green, 0; blue, 0 }  ][line width=0.08]  [draw opacity=0] (7.14,-3.43) -- (0,0) -- (7.14,3.43) -- cycle    ;
\draw    (317.73,246.58) -- (317.99,252.67) ;
\draw [shift={(318.12,255.67)}, rotate = 267.54] [fill={rgb, 255:red, 0; green, 0; blue, 0 }  ][line width=0.08]  [draw opacity=0] (7.14,-3.43) -- (0,0) -- (7.14,3.43) -- cycle    ;
\draw    (317.94,274.82) -- (318.18,280.77) -- (318.2,281.25) ;
\draw [shift={(318.33,284.25)}, rotate = 267.49] [fill={rgb, 255:red, 0; green, 0; blue, 0 }  ][line width=0.08]  [draw opacity=0] (7.14,-3.43) -- (0,0) -- (7.14,3.43) -- cycle    ;
\draw    (478.72,305.96) -- (487.3,286.37) ;
\draw [shift={(488.51,283.62)}, rotate = 113.67] [fill={rgb, 255:red, 0; green, 0; blue, 0 }  ][line width=0.08]  [draw opacity=0] (7.14,-3.43) -- (0,0) -- (7.14,3.43) -- cycle    ;
\draw    (478.72,305.96) -- (486.97,324.39) ;
\draw [shift={(488.19,327.13)}, rotate = 245.89] [fill={rgb, 255:red, 0; green, 0; blue, 0 }  ][line width=0.08]  [draw opacity=0] (7.14,-3.43) -- (0,0) -- (7.14,3.43) -- cycle    ;
\draw    (508.24,276.09) -- (515.12,276.05) ;
\draw [shift={(518.12,276.04)}, rotate = 179.69] [fill={rgb, 255:red, 0; green, 0; blue, 0 }  ][line width=0.08]  [draw opacity=0] (7.14,-3.43) -- (0,0) -- (7.14,3.43) -- cycle    ;
\draw    (509.9,319.58) -- (516.78,319.55) ;
\draw [shift={(519.78,319.53)}, rotate = 179.69] [fill={rgb, 255:red, 0; green, 0; blue, 0 }  ][line width=0.08]  [draw opacity=0] (7.14,-3.43) -- (0,0) -- (7.14,3.43) -- cycle    ;
\draw    (347.66,292.7) -- (353.05,292.67) ;
\draw [shift={(356.05,292.66)}, rotate = 179.69] [fill={rgb, 255:red, 0; green, 0; blue, 0 }  ][line width=0.08]  [draw opacity=0] (7.14,-3.43) -- (0,0) -- (7.14,3.43) -- cycle    ;
\draw    (393.08,293.57) -- (398.48,293.54) ;
\draw [shift={(401.48,293.52)}, rotate = 179.69] [fill={rgb, 255:red, 0; green, 0; blue, 0 }  ][line width=0.08]  [draw opacity=0] (7.14,-3.43) -- (0,0) -- (7.14,3.43) -- cycle    ;

\draw (300,231) node [anchor=north west][inner sep=0.75pt]   [align=left] {{\scriptsize Pooling}};
\draw (300,261) node [anchor=north west][inner sep=0.75pt]   [align=left] {{\scriptsize Masking}};
\draw (378,260) node [anchor=north west][inner sep=0.75pt]  [rotate=-90] [align=left] {{\scriptsize Transformer}};
\draw (423.13,247.7) node [anchor=north west][inner sep=0.75pt]   [align=left] {Encoder};
\draw (474,292) node [anchor=north west][inner sep=0.75pt]  [rotate=-90] [align=left] {{\scriptsize Fusion}};
\draw (504,255) node [anchor=north west][inner sep=0.75pt]  [rotate=-90] [align=left] {{\scriptsize FC Mean}};
\draw (505,310) node [anchor=north west][inner sep=0.75pt]  [rotate=-90] [align=left] {{\scriptsize FC Std}};
\draw (235,239.5) node [anchor=north west][inner sep=0.75pt]   [align=left] {{\scriptsize Word2Vec}};
\draw (246,276.4) node [anchor=north west][inner sep=0.75pt]   [align=left] {{\scriptsize Bbox}};
\draw (523,260) node [anchor=north west][inner sep=0.75pt]   [align=left] {{\scriptsize Mean}};
\draw (525,304) node [anchor=north west][inner sep=0.75pt]   [align=left] {{\scriptsize Std}};
\draw (302,289) node [anchor=north west][inner sep=0.75pt]   [align=left] {{\scriptsize Vis. Feat}};
\draw (408,290) node [anchor=north west][inner sep=0.75pt]   [align=left] {{\scriptsize Vis. Feat}};
\draw (390,200) node [anchor=north west][inner sep=0.75pt]   [align=left] {{\scriptsize Image Feature Map}};
\draw (379,242) node [anchor=north west][inner sep=0.75pt]   [align=left] {{\scriptsize K,V}};
\draw (348,295) node [anchor=north west][inner sep=0.75pt]   [align=left] {{\scriptsize Q}};

\end{tikzpicture}
\caption{Detailed encoder architecture.}
\label{fig:encoder}
\end{figure}
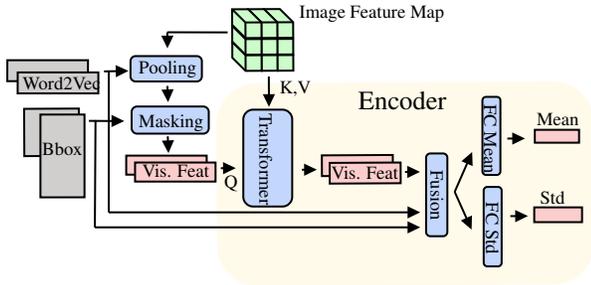

\section{Method}
For the rest of the paper we will refer to the pair $<s,o>$ as the context of a predicate $p$ in a triplet $<s,p,o>$. In similar fashion with ~\cite{Lu_2016} we aim to create a method, we name it RelVAE, that produces similar features for similar contexts. Given that most predicate's contexts exhibits a similarity and that RelVAE produces similar features for these contexts, our model may be able to generalize to unseen triplets. In order to achieve this functionality we use a conditional VAE that is trained on a set of contexts and use the features on the latent space of the model as the context representation. In the following we firstly describe the pretraining method and then the details of the few-shot classification head.
\subsection{Pretraining the cVAE}
The set used for pretraining consists of $(I_i,B_{i,s},B_{i,o},L_{i,s},L_{i,o})$ tuples  where $I_i$ is the Image, $B_{i,s},B_{i,o}$ are the bounding boxes of the subject and object, and $L_{i,s},L_{i,o}$ are the corresponding object labels. The cVAE takes as input the pair of bounding boxes and the labels while conditioning on the image through a transformer. This choice was motivated by the idea that this way the model is able to encode mutlimodal information like spatial, semantic and visual cues in the latent space while avoiding to encode image-dependent visual information that is irrelevant when describing a relation(\eg in a $<man,sitting-on,chair>$ relation, the color of the shirt or chair is of no importance to us). We argue that this characteristic of the cVAE's latent space makes it great feature extractor for the contexts. In the following we describe our motivation and choices to end up with the architecture show in Fig~\ref{fig:encoder}.

\textbf{Architecture.}
As image features we use the output of a resnet50~\cite{He_2015} before the Average Pooling Layer. We choose to give as input to the model the pair of semantics and bounding boxes of the subject and object. To model the semantics we use the corresponding word2vec~\cite{Mikolov_2013} embedding for each label. Using these embeddings and the image features we perform an attentional pooling to get two semantically dependent image feature maps, one for each entity in the relationship. Later the encoder uses the given bounding boxes to mask and extract the corresponding feature for the subject and object locations. We treat these two features as a sequence of two. We 
feed this sequence as input to a transformer where the queries are extracted from the above sequence while the keys and values are extracted from the sequence of image patches. The transformer is modified as to return the weighted average of the original image features(\ie instead of using the network that estimates the $values$ we use the identity function). We get as a result enriched visual features. At last we compute spatial features derived by the coordinates of the bounding boxes. We concatenate the visual, semantic and spatial features and fuse them with a feed forward network. The resulting vector is employed to estimate the mean and variance of the posterior distribution. 

The decoder is similar to the encoder. During the decoding phase, the sampled latent vector is given as input to two fully connected networks that estimate the subject and object word2vec embeddings without conditioning on the image. Furthermore two additional fully connected networks estimate the corresponding visual features. These features are treated in a similar way as in the encoding phase: they are fed into a modified transformer layer where the queries are extracted from them while the keys and values are extracted from the sequence of image patches. We choose not to predict directly the bounding boxes coordinates and justify this decision in the next paragraph. These image-conditioned visual features and the word2vec embeddings predicted above are the outputs of the model. 

\textbf{Losses.} Traditionally the  VAE loss cosists of the reconstruction loss and the KL divergence between the approximate posterior and the normal distribution. Our reconstruction loss consists  of 2 terms: a cosine distance loss between the ground truth $y_{i,sem}$ and predicted $\hat{y}_{i,sem}$ word2vec embeddings, namely $L_{cos}$, and a loss related to the bounding boxes $L_{bbox}$ described below. We choose the cosine distance instead of cross entropy to be able to generalize to object categories besides the ones in our training dataset. During testing we output the word2vec embedding that is the closest to the predicted with respect to the cosine distance. Regarding the $L_{bbox}$ loss, instead of predicting the bounding boxes coordinates from the sampled latent code we apply a binary cross entropy loss forcing the heatmaps of the decoder's transformer $H_i$ to match the binary masks derived from the bounding boxes $M_i$. Which choose this more difficult task to avoid overfitting on spatial data. We further apply a MSE loss on the g.t $y_{i,vis}$ and predicted visual features $\hat{y}_{i,vis}$ to enforce the model to encode visual cues. Formaly: 
\begin{equation}
    L_{cos} = \frac{1}{N}\sum_i^NCosineDistance(y_{i,sem},\hat{y}_{i,sem})
\end{equation}
\begin{equation}
    L_{bbox} = \frac{1}{N} \sum_i^N BCE(H_i,M_i)
\end{equation}
\begin{equation}
    L_{MSE} = \frac{1}{N} \sum_i^N \lVert y_{i,vis}- \hat{y}_{i,vis} \rVert_2^2
\end{equation}
\begin{equation}
    L_{total} = \alpha L_{cos} + L_{MSE} + L_{bbox} + \beta L_{KL}
\end{equation}
 where $N$ is the batch size and $\alpha,\beta$ are weighting coefficients.

\subsection{Few-shot classifiers}
The desired product of the above training process is an Encoder that given a context produces it's feature in a meaningful latent space, able to encode semantic,visual and spatial information. We exploit these properties by training simple predicate classifiers using this cVAE as a backbone. Specifically we freeze the cVAE, for each relation we use the output of the encoder as the context representation and treat the problem as a simple classification task. We then employ a feed-forward network with three layers and apply the cross entropy loss between the ground truth and the predicted predicates.



\begin{table*}
\begin{center}
\begin{tabular}{|c|c|c|c|}
\hline
Model & $R@50, k=1$ & $R@50,k=2$ & $R@50,k=5$ \\
\hline\hline
 LS & $4.51 \pm 1.70$ & $6.96 \pm 6.18$ & $13.02 \pm 3.60$ \\
V & $1.44 \pm 1.01$ & $6.35 \pm 5.32$ & $1.88 \pm 1.17$ \\
VLS & $2.79 \pm 1.41$ & $5.63 \pm 3.60$ & $9.16 \pm 2.34$ \\
Motifs~\cite{Zellers_2017} & $0.88 \pm 1.57$ & $0.70 \pm 0.98$ & $1.39 \pm 1.99$ \\
ATR-Net~\cite{Gkanatsios_2019b} & $2.91 \pm 2.55$ & $1.04 \pm 1.14$ & $3.37 +- 3.46$\\
relVAE+clf& $\bold{5.68 \pm 0.57}$ & $\bold{9.65 \pm 0.72}
$ & $\bold{13.38 \pm 0.31}$ \\
\hline
relVAE + KNN-1 & $6.88$ & $10.59$ & $11.08
$\\
\hline
\end{tabular}
\end{center}
\caption{Results on the VG200 dataset. We show the average and standard deviation of five random initializations. We use the baselines defined in~\cite{Gkanatsios_2020} where LS: Language and Spatial, V:Visual and VLS:Visual,Language and Spatial.}
\label{VG200 results}
\end{table*}

\begin{table}
\begin{center}
\begin{tabular}{|c|c|c|}
\hline
Model & $R@50, k=1$ & $R@50,k=2$\\
\hline\hline
 LS & $9.08 \pm 2.06$ & $12.00 \pm 3.82$ \\
V & $3.48 \pm 1.05$ & $4.72 \pm 2.30$ \\
VLS & $3.50 \pm 1.46$ & $4.72 \pm 1.20$ \\
Motifs~\cite{Zellers_2017} & $0.72 \pm 0.84$ & $2.13 \pm 2.51$ \\
ATR-Net~\cite{Gkanatsios_2019b} & $3.28 \pm 2.62$ & $3.81 \pm 3.29$ \\
relVAE+clf& $\bold{9.17 \pm 0.29}$ & $\bold{13.01 \pm 0.1}
$ \\
\hline
relVAE + KNN-1 & $9.52$ & $11.54$\\
\hline
\end{tabular}
\end{center}
\caption{Results on the VRD dataset. We show the average and standard deviation of five random initializations.}
\label{VRD results}
\end{table}

\section{Experiments}
We perform a number of quantitative and qualitative experiments to evaluate the power and efficiency of the model. Initially we compare our model with various baselines on the task of predicate classification and later we describe qualitative experiments that attempt to examine the properties of the latent space.

\textbf{Datasets:} We pretrain the cVAE on the VG200 dataset, a popular subset of the Visual Genome dataset~\cite{Krishna_2017} containing 150 object categories and 50 predicate categories commonly used in the 
 literature~\cite{Yang18}. We attempt few-shot classification both on the VG200 dataset and the VRD dataset~\cite{Lu_2016}, a smaller dataset consisting of 100 object categories and 70 predicates categories. For the few-shot experiments we select a subset of the training split of VG200 and VRD datasets and attempt predicate classification on the whole test set. More specifically we select $k$ examples for each predicate with $k \in \{1,2,5\}$ for the VG200 and $k \in \{1,2\}$ for the VRD. All few-shot splits will be made available.

 \textbf{Evaluation:} There are four common evaluation protocols with respect to the information given as input:
 \begin{itemize}
     \item \textbf{PredDet:} Given ground truth bounding boxes, object labels and which objects interact predict the interaction type.
     \item \textbf{PredCls:} Given ground truth bounding boxes and object labels predict which objects interact and the interaction type.
    \item \textbf{SgCls:} Given ground truth bounding boxes predict the object labels, which objects interact and the interaction type.
    \item \textbf{SgGen:} Given an image detect the object locations, predict the object labels, which of them interact and the interaction type.
 \end{itemize}
 Herein we are concerned with the $PredDet$ task. RelVAE can be extended to the $PredCls$ task by incorporating a pair filter and to the more complex task of $SgCls$, $SgGen$ by incorporating an object detector. Although adding the noise of object detector and pair filtering to the predictions will limit our conclusions about the performance of RelVAE since our goal is to examine if it is possible to capture the variation 
 in the context of relations and exploit it. Thus we leave this extension to future work. Our evaluation metric is $Recall@k$~\cite{Lu_2016} $with-graph-constraints$ that measures the fraction of the ground truth relations contained on the top-k predictions, allowing only one prediction per 
 pair.

\begin{table*}
\begin{center}
\begin{tabular}{|c|c|c|c|}
\hline
Model & $R@50, k=1$ & $R@50,k=2$ & $R@50, k=5$\\
\hline\hline
$~\cite{Dornadula_2019}^{+}$ & $12.00$ & $15.00$ & $20.00$ \\
relVAE+clf& $\bold{20.32 \pm 0.67}$ & $\bold{28.86\pm1.92}$ & $\bold{36.56\pm0.59}$ \\
\hline
\end{tabular}
\end{center}
\caption{Results on the examples of the test set of the VG200 dataset containing only the 25 least common predicates. We show the average and standard deviation of five random initializations. ($^+$ The authors of ~\cite{Dornadula_2019} do not provide numerical results, instead they show the results on plots. Here we give an approximation on their performance from this plot and refer the reader to Figure~4 of  ~\cite{Dornadula_2019}).}
\label{rare preds results}
\end{table*}

\begin{figure*}[t]
\begin{center}
\includegraphics[width=1\linewidth]{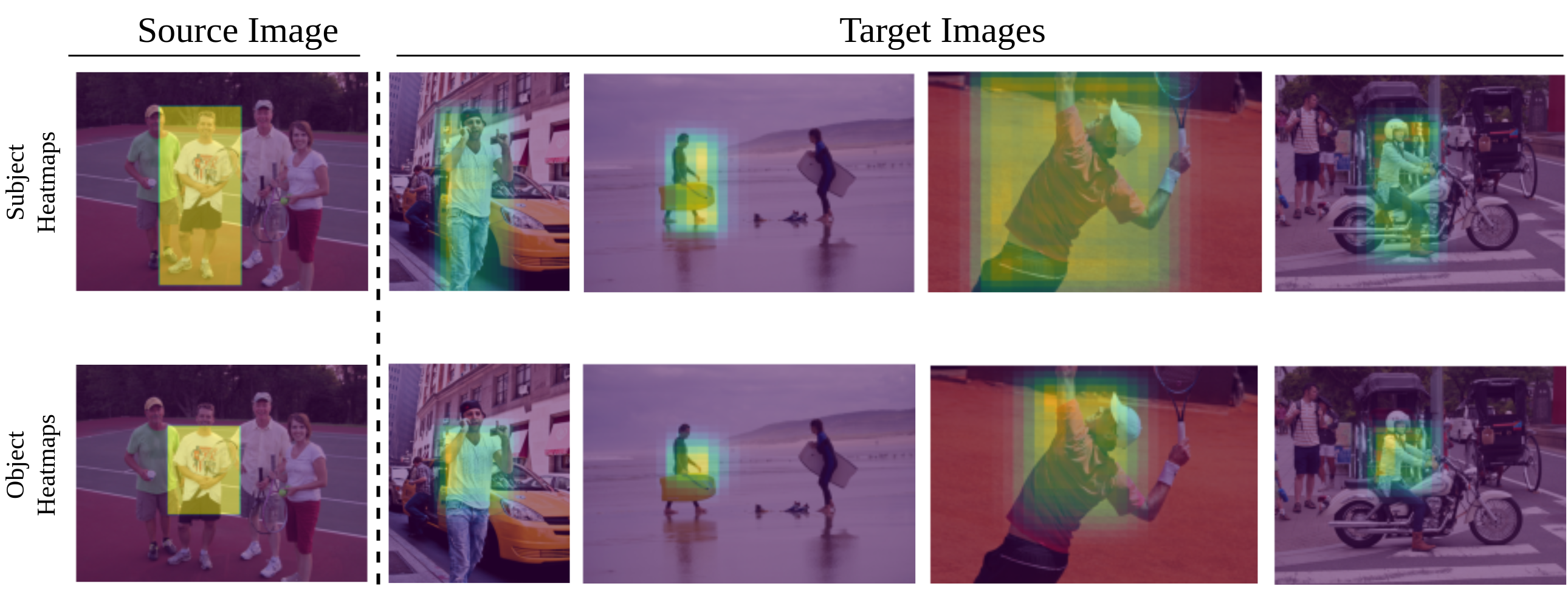}
\end{center}
   \caption{The above "cross reconstruction" experiment showcases the ability of the model to encode on the latent space semantic and visual information. More specifically we decode a $<man,shirt>$ context from a source Image on five different target Images. First and second row show the heatmap of the subject and object respectively.}
\label{fig:long}
\label{fig:onecol}
\label{fig:cross-reconstruct}
\end{figure*}

\subsection{Few-shot experiments}

To exploit the aforementioned abilities of the model we deal with  
the predicate classification problem on VG200 and VRD under the few-shot regime. Training on such a limited number of examples (the biggest dataset contains only 250 samples corresponding to 5 examples for each of the 50 predicates of the VG200 dataset) without some kind of pretraining is impossible and since to the best of our knowledge we are the first to introduce such a pretraining scheme, comparing against models of the literature that require more data gives us an unfair advantage. For this reason we implement a number of baselines with approximately the same number of parameters as our classification head. In a similar way with ~\cite{Dornadula_2019} we also include a popular model ~\cite{Zellers_2017} and a more recent model ~\cite{Gkanatsios_2019b} for completeness. The results are shown on Table~\ref{VG200 results} and Table~\ref{VRD results}. We use the three baselines as defined in ~\cite{Gkanatsios_2020}: 1)~LS, that uses only semantic and spatial information, 2)~V, that uses only visual information and 3)~VLS, that uses visual,semantic and spatial information. 
Surpassing all the baselines we achieve both higher $Recall@50$ and lower standard deviation across the five runs on both datasets. Finally, we also highlight that the baselines are highly sensitive to the initialization.

In ~\cite{Dornadula_2019} the authors first pretrain their model using the examples of the 25 most common predicates and later use their model as feature extractor to train k-shot classifiers for the 25 least common predicates. Even though we pretrain our model \emph{without} predicates at all we train k-shot classifiers for the same 25 least common predicates with $k\in \{1,2,5\}$. Our results are shown in Table~\ref{rare preds results}. We outperform their model by at least $8$ $ R@50$ on $k=1$, $13$ $R@50$ on $k=2$ and $16$ $R@50$ on $k=5$. We cannot be sure for the exact margin since in their work they show the results on a plot. Their results for $k=1,2,5$ are approximately $12,15,20$ $R@50$ respectively, we refer the reader to figure 4 of ~\cite{Dornadula_2019} for more details.

\begin{figure}[t]
\centering
   \begin{subfigure}[b]{0.45\textwidth}
   \includegraphics[width=\textwidth]{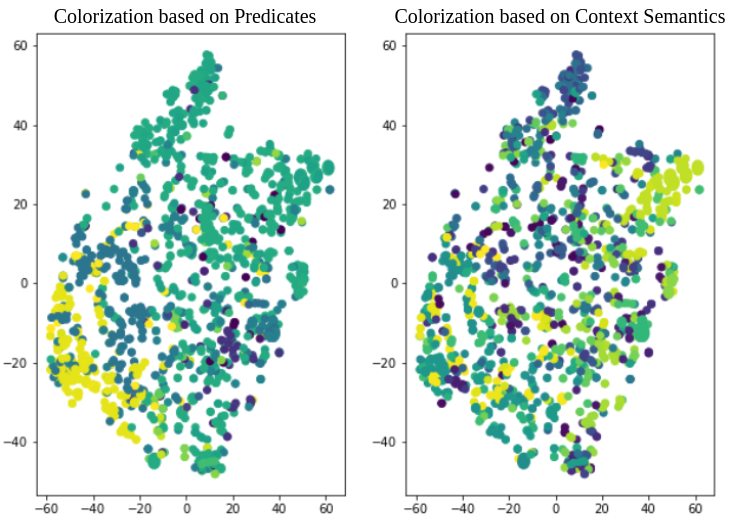}
   \caption{}
   \label{fig:TSNE1} 
\end{subfigure}
\begin{subfigure}[b]{0.45\textwidth}
   \includegraphics[width=\textwidth]{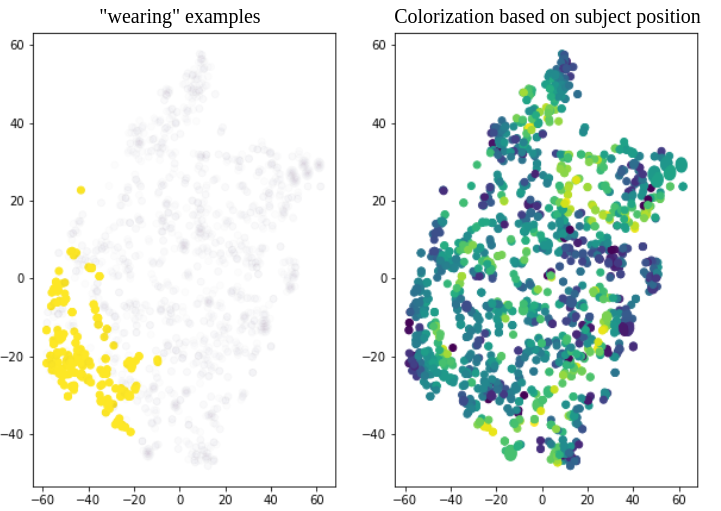}
   \caption{}
   \label{fig:TSNE2}
\end{subfigure}
\caption{(a) On the left plot the points are colorized based on the ground truth predicate of the relationship, on the right based on the semantics of the context. (b) On the left plot we highlight the region corresponding to the predicate "wearing" and on the right we colorize the points with respect to the normalized position of the subject inside the image. See the text for more details.}
\end{figure}

\subsection{Latent space properties}
Herein, we perform qualitative experiments to examine the ability of the model to capture semantic and spatial information.
Firstly, we inspect the latent space using dimensionality reduction algorithms and coloring the points with various criteria. Later we investigate encoded information by encoding a context from a source image $I_s$ and decoding the produced latent vector on a target image $I_t$. To evaluate the success of the experiment we look at the heatmaps of the decoder's transformer that indicates similarity between the latent code and the patches of the image.

\subsubsection{Latent space inspection}
Our goal is to approximately  visualize the latent codes (the features of contexts) produced by the encoder after feeding a subset of the testing dataset into the VAE. We use the t-SNE~\cite{Maaten2008} algorithm to reduce the latent code dimensionality down to 2 and colorize the 2d points using various criteria. In Fig.~\ref{fig:TSNE1} we provide two colorizations. On the left the latent space has been colorized using the predicate that corresponds to each point while on the right the points are colorized depending on the semantics of subject and object. We emphasize that during training and inference no predicate annotations have been used, although we see that a rough clustering on points of the same predicate occured on the first plot. Regarding the second colorization we see that colors in various places are blended. This is the case since there are multiple contexts that have different semantics, so different color, but are semantically and spatial similar(\eg contexts such as $<man,shirt>$, $<girl,shirt>$ and $<man,jacket>$ are all located on the bottom left part of the scatter plot). We highlight this specific location on Fig.~\ref{fig:TSNE2} on the left plot containing points belonging to predicate "wearing". 

Besides the semantic similarity that we show that the model encodes we also investigate to what extent is the model capable to encode spatial information. To this end we colorize the output of t-SNE based on the normalized horizontal position of the bounding box of the subject inside the image. As we show in Fig.~\ref{fig:TSNE2} on the right, inside semantically similar clusters (\eg the bottom left shown before) there is a grading regarding the spatial position producing this color gradient. These experiments suggest that the latent space maybe is "partitioned" based on the semantics while inside each partition points that are close between them belong to spatially similar instances of the context.

\subsubsection{The role of semantic information} To further evaluate the ability of the model to encode semantic information we 
choose to encode and decode a context that is present in both a source and target image $I_s,I_t$ respectively. In Fig.~\ref{fig:cross-reconstruct} we decode a $<man,shirt>$ from a source image to four different target images. The heatmaps suggest that the model is able to encode the semantic information since it correctly focuses on the given subject-object pair while the predicted word2vec embeddings match the ground truth.
This showcases that the latent space is able to capture the visual and spatial variability of a relation, a key property for few-shot learning.

\begin{figure}[t]
\begin{center}
\includegraphics[width=0.45\textwidth]{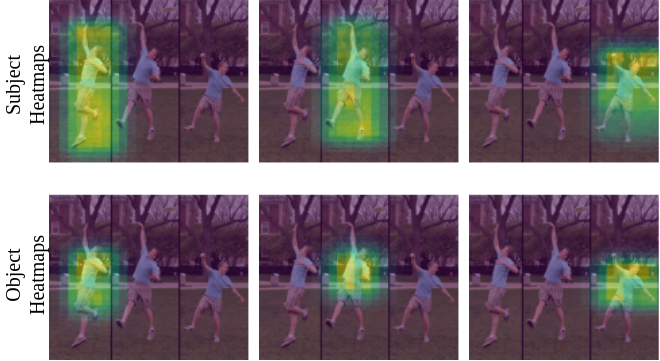}
\end{center}
   \caption{In this experiment we show the ability of the model to encode relative spatial information. We decode three latent vectors corresponding to three $<man,shirt>$ contexts on the left, center and right of the image from \emph{three different} source Images on a target Image containing three or more people to examine the focused location of the model.First and second row show the heatmap of the subject and object respectively.}
\label{fig:cross-reconstruct-lcr}
\end{figure}

\subsubsection{The role of spatial information} To examine the ability of the model on spatial information encoding we repeat the experiment with a different setting: This time we encode the same context in different locations (\eg left,center,right)  from different source images. We select a target image that contains this context in multiple locations, decode the different latent codes and examine which of the available locations in the target image the model chooses to focus. As it is shown in Fig~\ref{fig:cross-reconstruct-lcr}, the model each time focuses on the location encoded by the source image. This suggests that the model is capable of encoding relative spatial information.



\begin{figure}[t]
\centering
\includegraphics[width=1\linewidth]{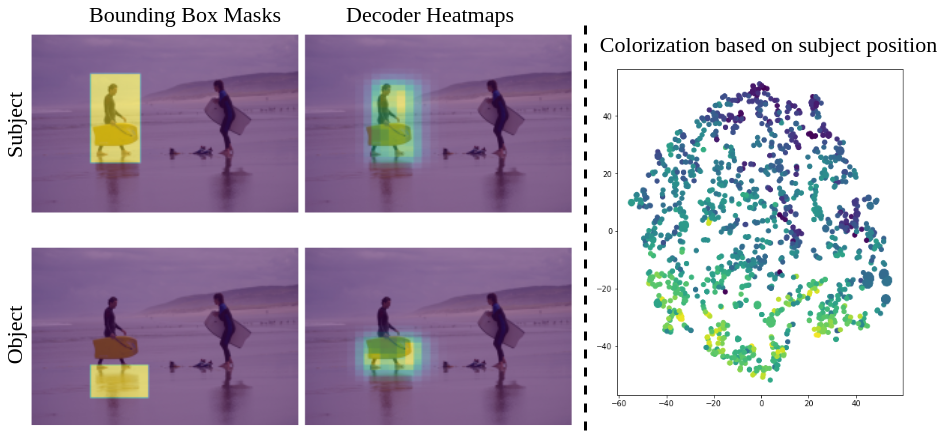}
\caption{\textbf{Left:} In this experiment we give as input a modified $<man,board>$ context where the semantics do not correspond to the location indicated by the bounding boxes, shown with yellow masks on the left image pair, and investigate which modality will prevail by examining the heatmaps of the decoded context, shown on the right image pair. \textbf{Right:} Points produced by an ablation on VAE colorized by subject position. More details on text.}
\label{fig:disrupt-vaecos}
\end{figure}

\begin{figure*}
\begin{center}
\scalebox{0.8}{\input{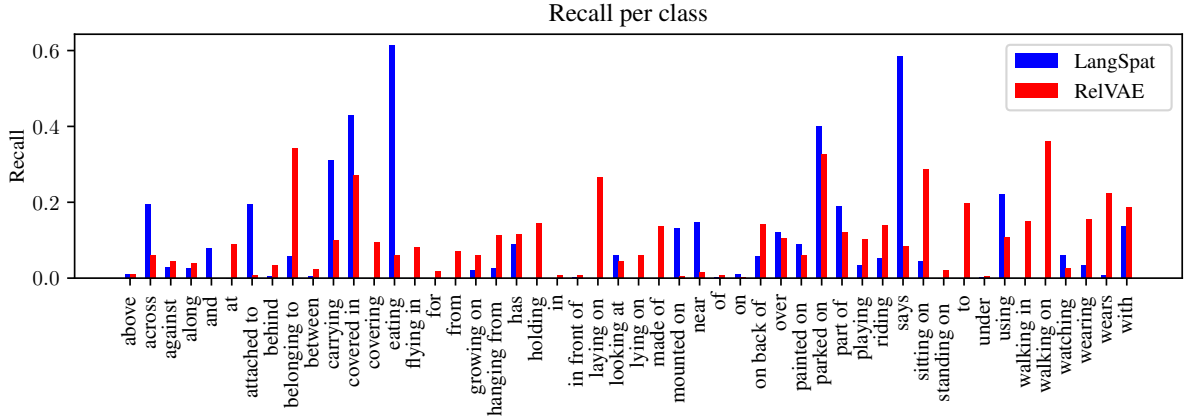}}
\end{center}
   \caption{We compare our model with the best performing baseline in terms of per-predicate recall. While they achieve comparable performances our model achieves a more balanced prediction distribution.}
\label{fig:short}
\label{fig:onecol}
\label{fig:recalls}
\end{figure*}


\subsection{Error analysis}
Given the fact that our model encodes both semantic and spatial information we are interested to see which of the two modalities dominates. To this end we encode a context where the semantics purposefully do not correspond with the location indicated by the input bounding boxes, as shown in fig~\ref{fig:disrupt-vaecos} on the left. We encode a context with semantics corresponding to $<man,board>$ while modifying the position of the input object bounding box to be closer to the feet and not on the surfboard. We observe that during decoding the heatmaps move towads the location described by the semantics and not the other way around. We attribute this to the independence on the image during the estimation of the word2vec embeddings. In contrast, we conduct an 
experiment where for the estimation of word2vec embeddings we use the features from the output of the transformer and show that in this new latent space the spatial information was more dominant. This is shown in Fig.~\ref{fig:disrupt-vaecos} on the right where we colorize the latent space of this new VAE with respect to the subject position in the image. In comparison with the right plot of Fig.~\ref{fig:TSNE2}, this time there is a color gradient across the whole space.

In an attempt to investigate the mispredictions of the model we examine the per-predicate recall on the VG200 dataset. We show a comparison plot between our model and the strongest baseline in Fig.~\ref{fig:recalls}. While the models achieve similar performances our model produces a more balanced prediction distribution in the sense that it achieves non zero recall for the 48 out of the 50 predicates in contrast with the 33 out of the 50 of the baseline. As we expected, our model achieves higher recall on predicates that occur with a specific type of contexts(\eg "wears" occurs with contexts such as $<man,shirt>$, $<woman,jacket>$ and so on) while it struggles with predicates without a specific type of context(\eg "in","and").

\section{Limitations and Future Work}%

In this work we presented a novel pretraining for VRD, RelVAE, that is able to encode semantic, visual and spatial information into a single latent space.
We also showed that the proposed pretraining enables few-shot predicate classification by conducting quantitative experiments both on VRD and VG200 datasets. Lastly we showcased the abilities of the latent space by performing various qualitative experiments.
Naturally the predicates that occur without a specific type of context need more examples in order to be understood. Learning to discriminate such predicates with only 1,2,5 examples is an extremely ill posed problem and as is shown on Fig.~\ref{fig:recalls} models fail to do so. By not taking as input the image, our classifier has not access to discriminative information that due to our "unsupervised pretraining" is not captured on the latent space, thus when increasing the data available during training more discriminative approaches ~\cite{Gkanatsios_2019b,Zellers_2017} achieve better performances. Regarding future work, RelVAE admits straightforward extensions to the tasks $PredCls$, $SgCls$ and $SgGen$, as mentioned earlier, by incorporating an object detector to detect and classify the objects and a pair filter to discriminate which objects interact. Another research direction is to extend the pretraining method to account for co-occurances by encoding not only a single context but the whole scene of contexts, most suited to the full data regime.


{\small
\bibliographystyle{ieee_fullname}
\bibliography{egbib}

\begin{thebibliography}{10}\itemsep=-1pt

\bibitem{Chen_2023_WACV}
Zhanwen Chen, Saed Rezayi, and Sheng Li.
\newblock More knowledge, less bias: Unbiasing scene graph generation with
  explicit ontological adjustment.
\newblock In {\em Proceedings of the IEEE/CVF Winter Conference on Applications
  of Computer Vision (WACV)}, pages 4023--4032, January 2023.

\bibitem{Dai_2017}
Bo Dai, Yuqi Zhang, and Dahua Lin.
\newblock {Detecting Visual Relationships with Deep Relational Networks}.
\newblock In {\em Proc. CVPR}, 2017.

\bibitem{Diomataris21}
Markos Diomataris, Nikolaos Gkanatsios, Vassilis Pitsikalis, and Petros
  Maragos.
\newblock Grounding consistency: Distilling spatial common sense for precise
  visual relationship detection.
\newblock In {\em 2021 IEEE/CVF International Conference on Computer Vision
  (ICCV)}, pages 15891--15900, 2021.

\bibitem{Dong_2022_CVPR}
Xingning Dong, Tian Gan, Xuemeng Song, Jianlong Wu, Yuan Cheng, and Liqiang
  Nie.
\newblock Stacked hybrid-attention and group collaborative learning for
  unbiased scene graph generation.
\newblock In {\em Proceedings of the IEEE/CVF Conference on Computer Vision and
  Pattern Recognition (CVPR)}, pages 19427--19436, June 2022.

\bibitem{Dornadula_2019}
Apoorva Dornadula, Austin Narcomey, Ranjay Krishna, Michael~S. Bernstein, and
  Li Fei-Fei.
\newblock {Visual Relationships as Functions: Enabling Few-Shot Scene Graph
  Prediction}.
\newblock In {\em Proc. ICCV Workshops}, 2019.

\bibitem{Gkanatsios_2019b}
Nikolaos Gkanatsios, Vassilis Pitsikalis, Petros Koutras, and Petros Maragos.
\newblock {Attention-Translation-Relation Network for Scalable Scene Graph
  Generation}.
\newblock In {\em Proc. ICCV Workshops}, 2019.

\bibitem{Gkanatsios_2020}
Nikolaos Gkanatsios, Vassilis Pitsikalis, and Petros Maragos.
\newblock {From Saturation to Zero-Shot Visual Relationship Detection Using
  Local Context}.
\newblock In {\em Proc. BMVC}, 2020.

\bibitem{Goel_2022_CVPR}
Arushi Goel, Basura Fernando, Frank Keller, and Hakan Bilen.
\newblock Not all relations are equal: Mining informative labels for scene
  graph generation.
\newblock In {\em Proceedings of the IEEE/CVF Conference on Computer Vision and
  Pattern Recognition (CVPR)}, pages 15596--15606, June 2022.

\bibitem{He_2015}
Kaiming He, Xiangyu Zhang, Shaoqing Ren, and Jian Sun.
\newblock {Deep Residual Learning for Image Recognition}.
\newblock {\em CoRR}, abs/1512.03385, 2015.

\bibitem{Kingma_2014}
Diederik~P. Kingma and Jimmy Ba.
\newblock {Adam: A Method for Stochastic Optimization}.
\newblock {\em CoRR}, abs/1412.6980, 2014.

\bibitem{Krishna_2017}
Ranjay Krishna, Yuke Zhu, Oliver Groth, Justin Johnson, Kenji Hata, Joshua
  Kravitz, Stephanie Chen, Yannis Kalantidis, Li-Jia Li, David~A. Shamma,
  Michael~S. Bernstein, and Li Fei-Fei.
\newblock Visual genome: Connecting language and vision using crowdsourced
  dense image annotations.
\newblock {\em Int. J. Comput. Vision}, 123(1):32–73, may 2017.

\bibitem{Li_2022_CVPR_2}
Lin Li, Long Chen, Yifeng Huang, Zhimeng Zhang, Songyang Zhang, and Jun Xiao.
\newblock The devil is in the labels: Noisy label correction for robust scene
  graph generation.
\newblock In {\em Proceedings of the IEEE/CVF Conference on Computer Vision and
  Pattern Recognition (CVPR)}, pages 18869--18878, June 2022.

\bibitem{Li_2022_CVPR}
Wei Li, Haiwei Zhang, Qijie Bai, Guoqing Zhao, Ning Jiang, and Xiaojie Yuan.
\newblock Ppdl: Predicate probability distribution based loss for unbiased
  scene graph generation.
\newblock In {\em Proceedings of the IEEE/CVF Conference on Computer Vision and
  Pattern Recognition (CVPR)}, pages 19447--19456, June 2022.

\bibitem{Lu_2016}
Cewu Lu, Ranjay Krishna, Michael Bernstein, and Li Fei-Fei.
\newblock {Visual Relationship Detection with Language Priors}.
\newblock In {\em Proc. ECCV}, 2016.

\bibitem{Lu21}
Yichao Lu, Himanshu Rai, Jason Chang, Boris Knyazev, Guangwei Yu, Shashank
  Shekhar, Graham~W. Taylor, and Maksims Volkovs.
\newblock Context-aware scene graph generation with seq2seq transformers.
\newblock In {\em 2021 IEEE/CVF International Conference on Computer Vision
  (ICCV)}, pages 15911--15921, 2021.

\bibitem{Mikolov_2013}
Tomas Mikolov, Kai Chen, Gregory~S. Corrado, and Jeffrey Dean.
\newblock {Efficient Estimation of Word Representations in Vector Space}.
\newblock {\em CoRR}, abs/1301.3781, 2013.

\bibitem{parcalabescu_2022}
Letitia Parcalabescu, Michele Cafagna, Lilitta Muradjan, Anette Frank, Iacer
  Calixto, and Albert Gatt.
\newblock {VALSE}: A task-independent benchmark for vision and language models
  centered on linguistic phenomena.
\newblock In {\em Proceedings of the 60th Annual Meeting of the Association for
  Computational Linguistics (Volume 1: Long Papers)}, pages 8253--8280, Dublin,
  Ireland, May 2022. Association for Computational Linguistics.

\bibitem{radford2021}
Alec Radford, Jong~Wook Kim, Chris Hallacy, Aditya Ramesh, Gabriel Goh,
  Sandhini Agarwal, Girish Sastry, Amanda Askell, Pamela Mishkin, Jack Clark,
  et~al.
\newblock Learning transferable visual models from natural language
  supervision.
\newblock In {\em International conference on machine learning}, pages
  8748--8763. PMLR, 2021.

\bibitem{rezende14}
Danilo~Jimenez Rezende, Shakir Mohamed, and Daan Wierstra.
\newblock Stochastic backpropagation and approximate inference in deep
  generative models.
\newblock In Eric~P. Xing and Tony Jebara, editors, {\em Proceedings of the
  31st International Conference on Machine Learning}, volume~32 of {\em
  Proceedings of Machine Learning Research}, pages 1278--1286, Bejing, China,
  22--24 Jun 2014. PMLR.

\bibitem{Sadeghi11}
Mohammad~Amin Sadeghi and Ali Farhadi.
\newblock Recognition using visual phrases.
\newblock In {\em CVPR 2011}, pages 1745--1752, 2011.

\bibitem{Snell2017}
Jake Snell, Kevin Swersky, and Richard Zemel.
\newblock Prototypical networks for few-shot learning.
\newblock In {\em Proceedings of the 31st International Conference on Neural
  Information Processing Systems}, NIPS'17, page 4080–4090, Red Hook, NY,
  USA, 2017. Curran Associates Inc.

\bibitem{Suhail21}
M. Suhail, A. Mittal, B. Siddiquie, C. Broaddus, J. Eledath, G. Medioni, and L.
  Sigal.
\newblock Energy-based learning for scene graph generation.
\newblock In {\em 2021 IEEE/CVF Conference on Computer Vision and Pattern
  Recognition (CVPR)}, pages 13931--13940, Los Alamitos, CA, USA, jun 2021.
  IEEE Computer Society.

\bibitem{Tang_2020}
Kaihua Tang, Yulei Niu, Jianqiang Huang, Jiaxin Shi, and Hanwang Zhang.
\newblock {Unbiased Scene Graph Generation from Biased Training}.
\newblock In {\em Proc. CVPR}, 2020.

\bibitem{Tang_2019}
Kaihua Tang, Hanwang Zhang, Baoyuan Wu, Wenhan Luo, and Weiwei Liu.
\newblock {Learning to Compose Dynamic Tree Structures for Visual Contexts}.
\newblock In {\em Proc. CVPR}, 2019.

\bibitem{Maaten2008}
Laurens van~der Maaten and Geoffrey Hinton.
\newblock Visualizing data using t-sne.
\newblock {\em Journal of Machine Learning Research}, 9(86):2579--2605, 2008.

\bibitem{Vinyals2016}
Oriol Vinyals, Charles Blundell, Timothy Lillicrap, koray kavukcuoglu, and Daan
  Wierstra.
\newblock Matching networks for one shot learning.
\newblock In D. Lee, M. Sugiyama, U. Luxburg, I. Guyon, and R. Garnett,
  editors, {\em Advances in Neural Information Processing Systems}, volume~29.
  Curran Associates, Inc., 2016.

\bibitem{Wang20}
Weitao Wang, Meng Wang, Sen Wang, Guodong Long, Lina Yao, Guilin Qi, and Yang
  Chen.
\newblock One-shot learning for long-tail visual relation detection.
\newblock {\em Proceedings of the AAAI Conference on Artificial Intelligence},
  34(07):12225--12232, Apr. 2020.

\bibitem{Xu_2017}
Danfei Xu, Yuke Zhu, Christopher~Bongsoo Choy, and Li Fei-Fei.
\newblock {Scene Graph Generation by Iterative Message Passing}.
\newblock In {\em Proc. CVPR}, 2017.

\bibitem{Yang18}
Jianwei Yang, Jiasen Lu, Stefan Lee, Dhruv Batra, and Devi Parikh.
\newblock Graph r-cnn for scene graph generation.
\newblock In Vittorio Ferrari, Martial Hebert, Cristian Sminchisescu, and Yair
  Weiss, editors, {\em Computer Vision -- ECCV 2018}, pages 690--706, Cham,
  2018. Springer International Publishing.

\bibitem{Zellers_2017}
Rowan Zellers, Mark Yatskar, Sam Thomson, and Yejin Choi.
\newblock {Neural Motifs: Scene Graph Parsing with Global Context}.
\newblock In {\em Proc. CVPR}, 2018.

\bibitem{Zhuang_2017}
Bohan Zhuang, Lingqiao Liu, Chunhua Shen, and Ian~D. Reid.
\newblock {Towards Context-Aware Interaction Recognition for Visual
  Relationship Detection}.
\newblock In {\em Proc. ICCV}, 2017.

\end{thebibliography}
}

\end{document}